\documentclass{ifacconf}

\usepackage{graphicx}      
\usepackage{natbib}        
\usepackage{amsmath}
\usepackage{amsfonts}
\theoremstyle{definition}
\newtheorem{dfn}{Definition}
\usepackage{bm}
\begin{document}
\begin{frontmatter}

\title{Human--Centered Cooperative Control Coupling Autonomous and Haptic Shared Control via Control Barrier Function}

\author[First]{Eito Sato} 
\author[Second]{Takahiro Wada} 

\address[First]{Nara Institute of Science and Technology (e-mail: sato.eito.sd7@is.naist.jp).}
\address[Second]{Nara Institute of Science and Technology (e-mail: t.wada@is.naist.jp).}

\begin{abstract}                
Haptic shared control (HSC) is effective in teleoperation when full autonomy is limited by uncertainty or sensing constraints. However, autonomous control performance achieved by maximizing HSC strength is limited because the dynamics of the joystick and human arm affect the robot’s behavior. We propose a cooperative framework coupling a joystick-independent autonomous controller with HSC. A control barrier function ignores joystick inputs within a safe region determined by the human operator in real-time, while HSC is engaged otherwise. A pilot experiment on simulated tasks with tele-operated underwater robot in virtual environment demonstrated improved accuracy and reduced required time over conventional HSC.

\end{abstract}


\end{frontmatter}
{\small © 2025 the authors. This work has been accepted to IFAC for publication under a Creative Commons Licence CC-BY-NC-ND.}


\section{Introduction}
As teleoperation systems become more widespread, there is an increasing need for control frameworks that effectively leverage machine intelligence. In environments where fully autonomous control (F-AC), which directly controls the systems, is constrained by uncertain sensing and environmental variations, cooperative control between humans and machines becomes indispensable (\cite{Flemisch2016}). Many studies have specifically focused on shared control, in which a human and an autonomous system simultaneously execute control of the same task and which can be categorized as haptic shared control (HSC) or input-mixing shared control (IMSC) (\cite{abbink2010neuromuscular}).

HSC enables the human operators and autonomous controllers to communicate continuously through the input forces of the same input device(\cite{abbinkLoHA2012}).
For example, improved accuracy in uncertain environments and reduced the operator's workload for a remotely operated underwater vehicle have been reported (\cite{konishi}).
In HSC, the strength of the input forces which is generated by the automated system is important for the control accuracy and operators' workload.

Abbink et al. introduced the concept of the level of haptic authority (LoHA), which explicitly indicates the magnitude of the guidance force of a machine(\cite{abbinkLoHA2012}).
Several previous studies have investigated methods to make LoHA variable, and have shown that varying LoHA according to the surrounding environment reduces the operator's workload and improves task accuracy(\cite{ZwaanLoHA2019, MarcanoLoHA2020}).
In addition, methods that allow humans to change the LoHA have also been studied, in which a supplemental gripper, functioning as a separate human–machine interface to determine the LoHA, is added alongside the joystick (\cite{sato, yamamoto}). In these methods, the angle of the supplemental gripper determines the LoHA, while the system’s confidence in sensing and control is communicated to the human through the gripper via its torque (\cite{sato}) and angle (\cite{yamamoto}), thereby addressing the perceived environmental difficulty in teleoperation.
As mentioned earlier, one of the strengths of HSC is its ability to smoothly connect autonomous control (AC) through the input device and fully manual control by adjusting the LoHA (\cite{abbinkLoHA2012}).

However, when aiming to maximize the use of AC, two key challenges arise: (1) the performance of AC achieved by maximizing the LoHA in HSC is limited because the dynamics of the joystick and human arm affect the resulting joystick movement, thereby influencing the robot’s behavior; and (2) even when AC performs well, its behavior may not fully align with the operator’s intent (\cite{itoh2016hierarchical}).
These challenges highlight the need for methods that allow the human to explicitly define and continuously modify the operational scope of F-AC, which controls without an input device, as needed.

To address these issues, we propose a new framework for human–machine cooperative control in which F-AC and HSC are smoothly coupled. In particular, this study focuses on human-centric cooperative control, in which the human ultimately determines the degree to which F-AC is utilized. Specifically, (1) the human defines the operating region of the F-AC through a dedicated interface, inspired by the gripper mechanism (\cite{sato, yamamoto}) for human modification of LoHA, and (2) within this F-AC region, signals from the input device used in HSC do not affect the robot’s behavior, whereas HSC is activated otherwise. 

To seamlessly connect these two controllers and to prevent conflicts between F-AC and HSC including human control when both are active, this study adopts a control barrier function (CBF) approach. The CBF approach was originally developed to theoretically guarantee that the system state remains within a predefined safe region (\cite{AmesCBF, WielandCBF}). An application of CBF to human–machine cooperative control was presented by \cite{nakamura}, where the system disregarded human inputs and applied F-AC whenever the state approached the predefined safe region boundary. Inspired by this method, we extend it to smoothly connect F-AC and HSC in a human-centric manner. More specifically, the human defines, via CBF, the region in which only F-AC is applied and signals from HSC are disregarded, ensuring that the system state remains within this region. Once the state exits the CBF-defined region, the robot is directly controlled according to the human’s intent with haptic guidance from HSC.

In summary, the purpose of the present study is to build a new human-centric cooperative framework that connects F-AC and HSC smoothly via a time-varying CBF approach. A human-in-the-loop pilot experiment investigates the functionality of the proposed method through a path-following task using a teleoperated underwater robot in a virtual environment.

\section{Controller Design}
In this chapter, we briefly explain the original CBF-based human-assist control framework~(\cite{nakamura}) and then present the details of our proposed method.

\subsection{Human-assist control}
In this subsection, we introduce a time-varying CBF formulation.

We consider the following control-affine system:
\begin{equation}
    \dot{\bm{x}} = f(\bm{x}) + g(\bm{x})(\bm{u}_h + \bm{u}),
    \label{eq:1}
\end{equation}
where $\bm{x} \in \mathbb{R}^n$ is the system state, $\bm{u}\in \mathbb{R}^m$ is the control input, and $\bm{u}_h\in \mathbb{R}^m$ is the human input. The functions $f \in \mathbb{R}^n$ and $g \in \mathbb{R}^n \times \mathbb{R}^m$ are assumed to be locally Lipschitz, which ensures the existence and uniqueness of the system trajectories.
In the context of the teleoperation system considered, this assumption is reasonable because the dynamic or kinematic model is composed of physically smooth functions such as inertia, damping, and external forces.

Moreover, we introduce the following set $\mathcal{G} \subset \mathbb{R}^{n+1}$ as the graph of a safe set $X(t)$.
\begin{equation}
    \mathcal{G}(X) = \{(\bm{x}, t)\in \mathbb{R}^{n+1} |\bm{x}(t)\in X(t)\}.
    \label{eq:1.5}
\end{equation}

Then, the definition of the time-varying CBF was described as follows.

\begin{dfn}
\label{def1}
Consider the system \eqref{eq:1} and the graph $\mathcal{G}(X)$, where $X(t)$ is a safe set. If the following conditions are satisfied, a $C^1$ function $B:\mathbb{R}^n\times\mathbb{R}\rightarrow \mathbb{R}$ is a time-varying control barrier function.
\begin{enumerate}
    \item $B$ is a non-negative function and the set $\{\bm{x}|B(\bm{x}, t_0)\leq L\}$ is compact for any $L\in\mathbb{R}\geq0$ and any fixed time $t_0$.
    \item There exist constants $K>0$ and $C>0$ such that the following inequality is satisfied:
    \begin{equation}
    \begin{split}
        &\inf_{\bm{u}\in\mathbb{R}^m}{\dot{B}(\bm{x},\bm{u}_h,\bm{u},t)} \\
        &=\inf_{\bm{u}\in\mathbb{R}^m}{[L_f B(\bm{x},t)+L_g B(\bm{x},t)\{\bm{u}_h(t)+\bm{u}\}} \\
        &\ \ \ \ \ \ +\frac{\partial B}{\partial t}(\bm{x},t)] \\
        &<KB(\bm{x},t)+C,
    \end{split}
    \label{eq:2}
    \end{equation}
    where $L_f B$ and $L_g B$ are the Lie derivatives as follows.
    \begin{equation*}
        L_f B=\frac{\partial B}{\partial \bm{x}}f(\bm{x}), L_g B=\frac{\partial B}{\partial \bm{x}}g(\bm{x}).
    \end{equation*}
\end{enumerate}
\end{dfn}

If a function $B$ is time-varying CBF, the human-assist control input is defined as follows:
\begin{equation}
        \bm{u} =
        \begin{cases}
                -\frac{I(\bm{x},\bm{u}_h, t)-Q(\bm{x}, t)}{(L_g B)(L_gB)^\top}(L_gB)^\top  \ \ (I>Q)\\ 
                \ \ \ \ \ \ \ \ \ \ \ \ \ \ \ \ \ \ \ \ \ \ \bm{0}  \ \ \ \ \ \ \ \ \ \ \ \ \ \ \ \ \ (I\leq Q)
        \end{cases},
    \label{eq:3}
\end{equation}
where functions $I:\mathbb{R}^n\times\mathbb{R}^m\times\mathbb{R}\rightarrow\mathbb{R}$ and $Q:\mathbb{R}^n\times\mathbb{R}\rightarrow\mathbb{R}$ are defined by the following equations.
\begin{equation}
    \begin{split}
        I(\bm{x},\bm{u}_h, t)&=L_fB(\bm{x},t)+L_gB(\bm{x},t)\bm{u}_h(t)+\frac{\partial B}{\partial t}(\bm{x},t), \\
        Q(\bm{x},t)&=KB(\bm{x},t)+C.
    \end{split}
    \label{eq:4}
    \end{equation}

\begin{figure}[t]
\centering
\includegraphics[width=\linewidth]{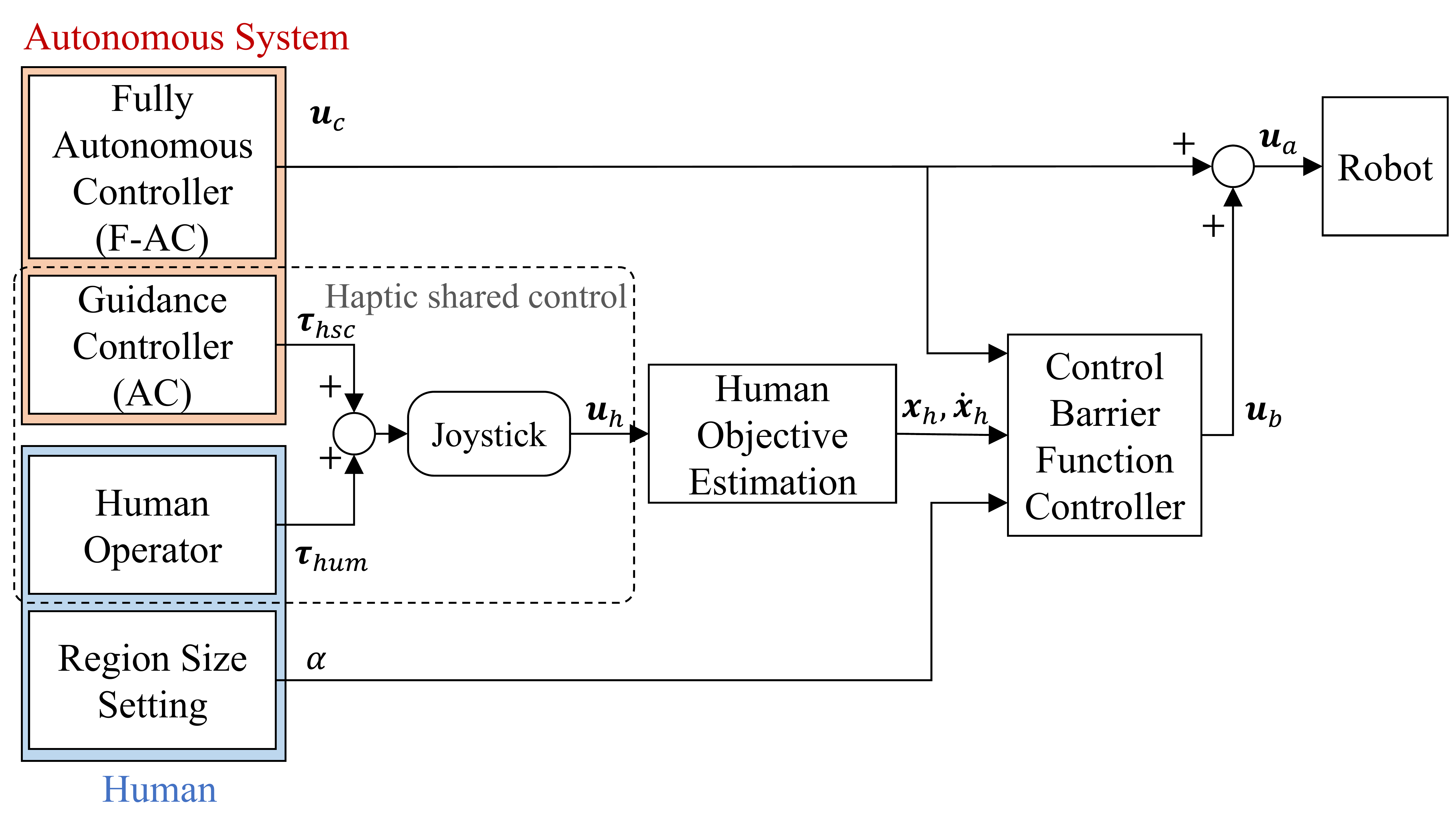}    
\vspace{-8mm}
\caption{The overall framework of the proposed method.} 
\vspace{5mm}
\label{fig:1}
\end{figure}

\subsection{Human-centered control barrier function approach}
We expand the previous method to human-centered control. An overall diagram of the proposed framework is shown in Fig.~\ref{fig:1}.

Now, we consider the following control-affine system:
\begin{equation}
    \dot{\bm{x}}_a = f(\bm{x}_a) + g(\bm{x}_a)(\bm{u}_c + \bm{u}_b),
    \label{eq:5}
\end{equation}
where $\bm{x}_a \in \mathbb{R}^n$ is the system state, $\bm{u}_c\in \mathbb{R}^m$ is the fully autonomous control (F-AC) input, and $\bm{u}_b\in \mathbb{R}^m$ is the control input. The functions $f$ and $g$ are assumed to be locally Lipschitz continuous.

Let the human-acceptable region be a circle, the center of which is $\bm{x}_h$ and the radius is $\alpha$, where $\alpha$ is adapted by the human operator.
That is, the candidate CBF function $B$ has the system state $\bm{x}_a$, the human desired position $\bm{x}_h$, and the acceptable region's radius $\alpha$ as arguments and is defined as follows.

\begin{equation}
B(\bm{x}_a,\bm{x}_h,\alpha)=
    \begin{cases}
    \frac{[(\bm{x}_h-\bm{x}_a)^\top(\bm{x}_h-\bm{x}_a)]^3}{l^2} \ &(l>0) \\
    (\bm{x}_h-\bm{x}_a)^\top(\bm{x}_h-\bm{x}_a) \ \ &(l\leq 0) 
    \end{cases}
    ,
\label{eq:6}
\end{equation}
where
\begin{equation}
l=\alpha^2-(\bm{x}_h-\bm{x}_a)^\top(\bm{x}_h-\bm{x}_a).
\label{eq:7}
\end{equation}
Note that $B$ is discontinuous at $\|\bm{x}_h-\bm{x}_a\|=\alpha$. In this case, the robot state within the acceptable region will not cross the boundary unless the decreasing speed of $\alpha$ exceeds the robot’s movement speed. This is necessary to define the out-of-region behavior because the state is in the acceptable region when $\|\bm{x}_h-\bm{x}_a\|<\alpha$ and out-of-region when $\|\bm{x}_h-\bm{x}_a\|>\alpha$. To make the behavior inside and outside the domain similar, the numerator is the third power when $l>0$.

Next, we redefine the functions $I$ and $Q$ in \eqref{eq:4} as
\begin{equation}
\begin{split}
    I(\bm{x}_a, \bm{u}_c, \bm{x}_h, \alpha)=&L_fB+L_gB\bm{u}_c+\frac{\partial B}{\partial \bm{x}_h}\bm{\dot{x}}_h, \\
    Q(\bm{x}_a, \bm{x}_h, \alpha)=&\text{sign}(l)\frac{\alpha}{\alpha_{max}}[KB(\bm{x}_a,\bm{x}_h,\alpha)+C] \\
    & \ -\frac{\partial B}{\partial \alpha}\dot{\alpha},
\end{split}
\label{eq:8}
\end{equation}
\noindent where $L_fB=\partial B/\partial \bm{x}_a f(\bm{x}_a)$ and $L_gB=\partial B/\partial \bm{x}_a g(\bm{x}_a)$ are the Lie derivatives.

In this case, the control input defined in the following equation allows both the position and size of the acceptable region to be specified by the human operator, while ensuring that the system state remains within that region.
\begin{equation}
    \bm{u}_b=
    \begin{cases}
    -\frac{I+J-Q}{(L_gB)(L_gB)^\top}(L_gB)^\top \ \ \ &(I>Q) \\
    \ \ \ \ \ \ \ \ \ \ \ \ \ \ \ \   \bm{0} &(I \leq Q)
    \end{cases}
    ,
\label{eq:9}
\end{equation}
where the function $J$ is described as follows;
\begin{equation}
    J = \max(0,-\text{sign}(l)\sqrt{I^2+\{(L_gB)^\top(L_gB)\}^2})
    \label{eq:10}.
\end{equation}
The function $J$ becomes active only when $l<0$, that is, when the system state lies outside the acceptable region. It was introduced to ensure smooth control input according to Sontag’s formula.

\begin{pf}
The derivative of human-centered CBF $B$ is calculated as follows:
\begin{equation}
\dot{B}=L_fB+L_gB(\bm{u}_c+\bm{u}_b)+\frac{\partial B}{\partial \bm{x}_h}\dot{\bm{x}}_h+\frac{\partial B}{\partial \alpha}\dot{\alpha}
\label{eq:11}.
\end{equation}

When $I \leq Q$, the following inequality holds according to \eqref{eq:8}:
\begin{equation}
\begin{split}
    \dot{B}&=I-Q+\text{sign}(l)\frac{\alpha}{\alpha_{max}}(KB+C) \\
    &\leq \text{sign}(l)\frac{\alpha}{\alpha_{max}}(KB+C)\\
    &\leq KB+C
    \label{eq:12}.
\end{split}
\end{equation}
This satisfies (2) of Definition 1.

On the other hand, $I>Q$, if the system state $\bm{x}_a$ is within the acceptable region, then the input by CBF is the same as in \eqref{eq:3}.
When $l<0$, i.e., the state is outside the region, the time derivative of $B$ is as follows.
\begin{equation}
    \begin{split}
    \dot{B}&=L_fB+L_gB\bm{u}_c+\frac{\partial B}{\partial \bm{x}_h}\dot{\bm{x}}_h+\frac{\partial B}{\partial \alpha}\dot{\alpha}-(I+J-Q)\\
    &=-\frac{\alpha}{\alpha_{max}}(KB+C)-\sqrt{I^2+\{(L_gB)^\top(L_gB)\}^2} \\
    &<KB+C.
    \end{split}
    \label{eq:13}
\end{equation}
Furthermore, \eqref{eq:13} guarantees that $\dot{B}<0$. When $\bm{x}_h\neq\bm{x}_a$, the state $\bm{x}_a$ converges to $\bm{x}_h$ because $B>0$.
Therefore, the control input \eqref{eq:10} converges the state to the human goal in the case where the state goes outside the region, such that the region becomes zero.
\end{pf}

\section{Pilot experiment}
A human-in-the-loop pilot experiment was conducted to investigate the operability and performance of the proposed method through a path-following task using a teleoperated underwater vehicle in a virtual environment. Four operators performed path-following tasks under both simple HSC and the proposed method, and their performance was compared in terms of tracking accuracy and task completion time.
This study was approved by the Research Ethics Committee of Nara Institute of Science and Technology (No. 2023-I-12).

\subsection{Scenario and settings}
\begin{figure}[t]
\centering
\includegraphics[width=\linewidth]{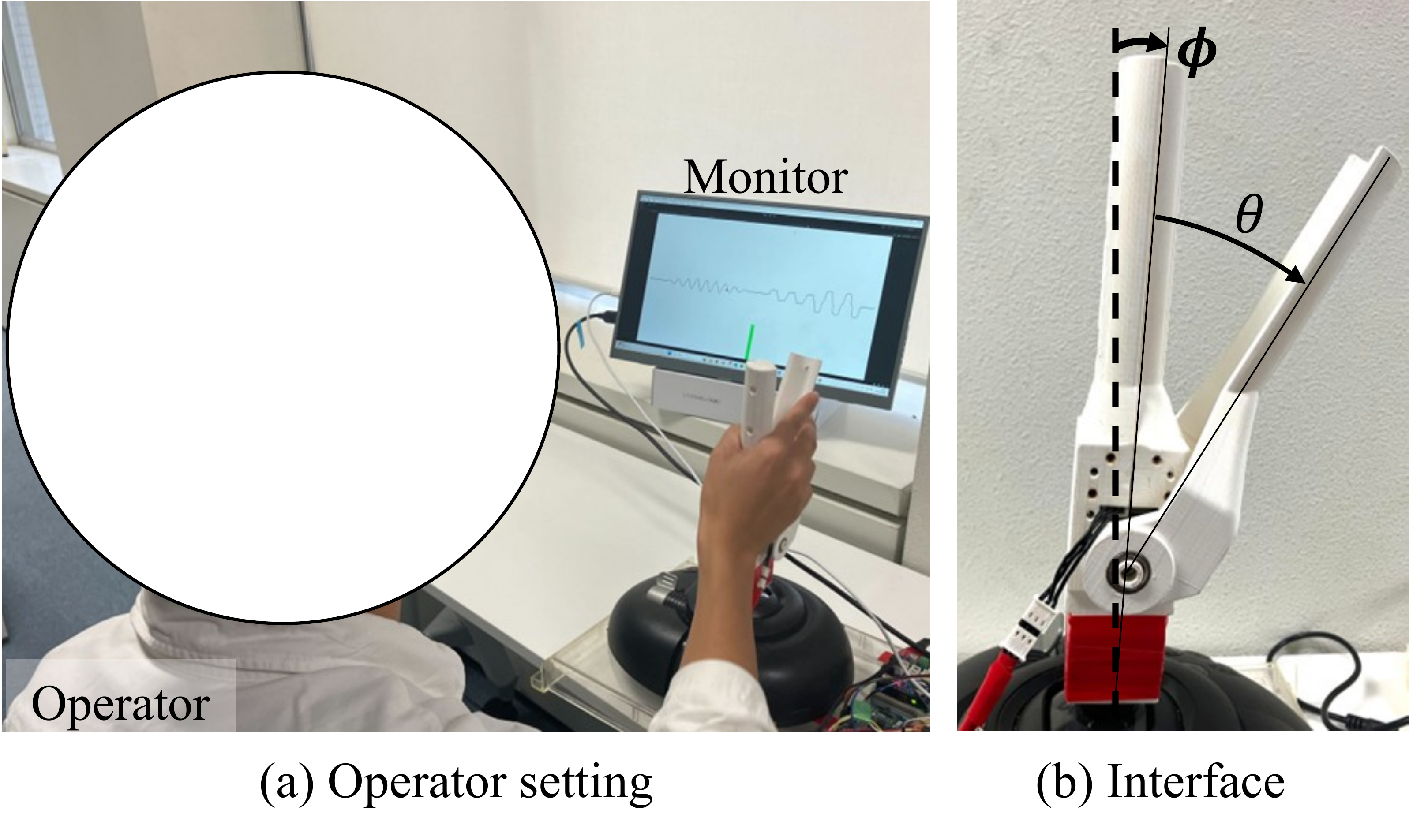}
\vspace{-5mm}
\caption{Simulation environment using Unity. The operator manipulates the joystick while monitoring a display.}
\label{fig:2}

\end{figure}
\begin{figure}[t]
\centering
\includegraphics[width=\linewidth]{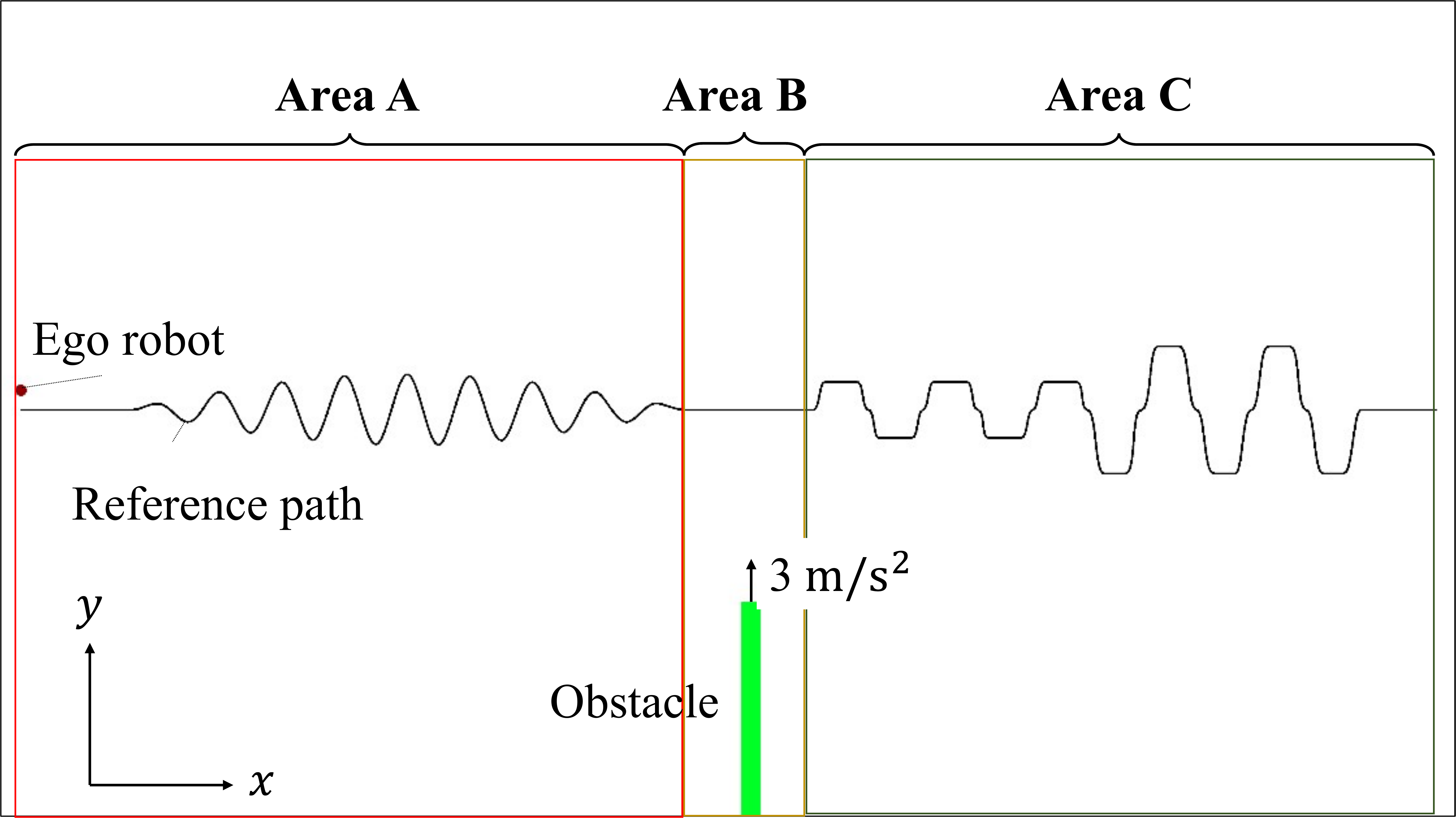}
\caption{Path environment of the path-following task in the simulation.}
\vspace{8mm}
\label{fig:3}
\end{figure}
The details are as follows.
The virtual underwater vehicle in the two-dimensional simulation environment was implemented in Unity, and the computation time of the control script in Unity was 2 ms. 
The operator follows a path using a flight joystick (SideWinder Force Feedback 2, Microsoft) while monitoring a display (Fig. \ref{fig:2}). 
As shown in Fig. \ref{fig:3}, the simulated path consists of three distinct regions: Area A, where smooth maneuvering is required; Area B, where human intervention is necessary to avoid collisions with obstacles (the green object shown in Fig. \ref{fig:3}) that moves upward at 3 m/s when the vehicle enters the area; and Area C, which demands high-precision control.
The maximum joystick tilt angles in the forward/backward and right/left directions were $20^\circ$ and $25^\circ$, respectively. Each motor driver operated at a maximum voltage of 6.0 V.
A gripper was attached to the joystick, and the gripper angle determined the desired radius of the acceptable region. 
The gripper motor (Dynamixel AX-12A, ROBOTIS) had a maximum torque of 0.7 Nm.

The HSC input $\bm{u}_h$ and the acceptable region radius $\alpha$ were obtained from the joystick and gripper angle according to the following equations:
\begin{equation}
    \begin{split}
        \bm{u}_h&=K_{joy}\bm{\phi}, \\
        \alpha &= K_{\theta}\theta.
    \end{split}
    \label{eq:14}
\end{equation}

In this simulation, a simple HSC is used for comparison with the proposed method. In both systems, the joystick moves automatically to generate autonomous input $\bm{u} _c$. 
Moreover, the time derivative of human desired position $\bm{\dot{x}}_h$ is defined as follows. 
\begin{equation}
    \bm{\dot{x}}_h=f(\bm{x}_h)+g(\bm{x}_h)\bm{u}_h.
\end{equation}
And for this simulation, we assume $f(\bm{x_a})=f(\bm{x}_h)=\bm{0}, g(\bm{x}_a)=g(\bm{x}_h)=I_{2\times2}$.  
Both $f$ and $g$ are local Lipschitz and satisfy the assumptions of the method.

Since it was difficult to obtain the time derivative of $\alpha$ in advance, a low-pass filter was used to obtain a new radius time derivative $\dot{\beta}$ as in the following equation.
\begin{equation}
    \dot{\beta}=(\alpha-\beta)/\epsilon.
\end{equation}

Furthermore, if the contraction speed of the acceptable region is greater than the maximum vehicle speed, a limit is placed on the contraction speed because the system state may sometimes leave the acceptable region.
\begin{equation}
\dot{\beta}_{max}=
    \begin{cases}
        2 + |\dot{\bm{x}}_h|  &((\bm{x}_h-\bm{x}_a)^\top\bm{x}_h>0)\\
        2 - |\dot{\bm{x}}_h|  &((\bm{x}_h-\bm{x}_a)^\top\bm{x}_h\leq0)
    \end{cases}
     .
\end{equation}
\subsection{Result}

\begin{figure}[t]
\centering
\includegraphics[width=\linewidth]{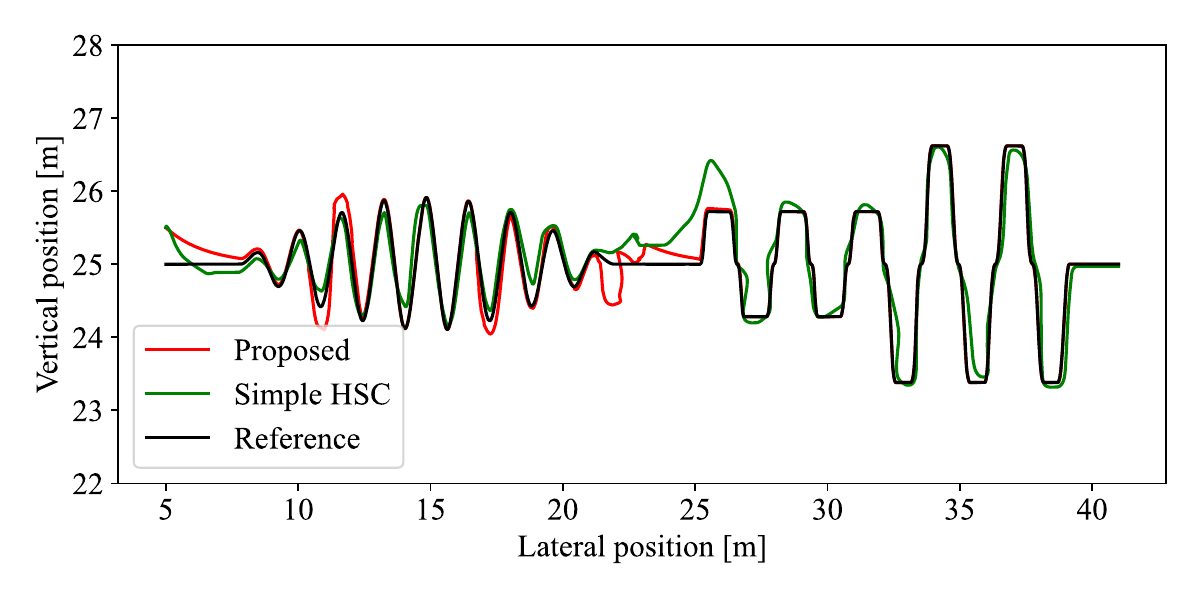}
\vspace{-8mm}
\caption{Trajectories of a representative operator using the simple HSC and proposed methods.}
\vspace{5mm}
\label{fig:4}
\end{figure}

\begin{figure}[t]
\centering
\includegraphics[width=\linewidth]{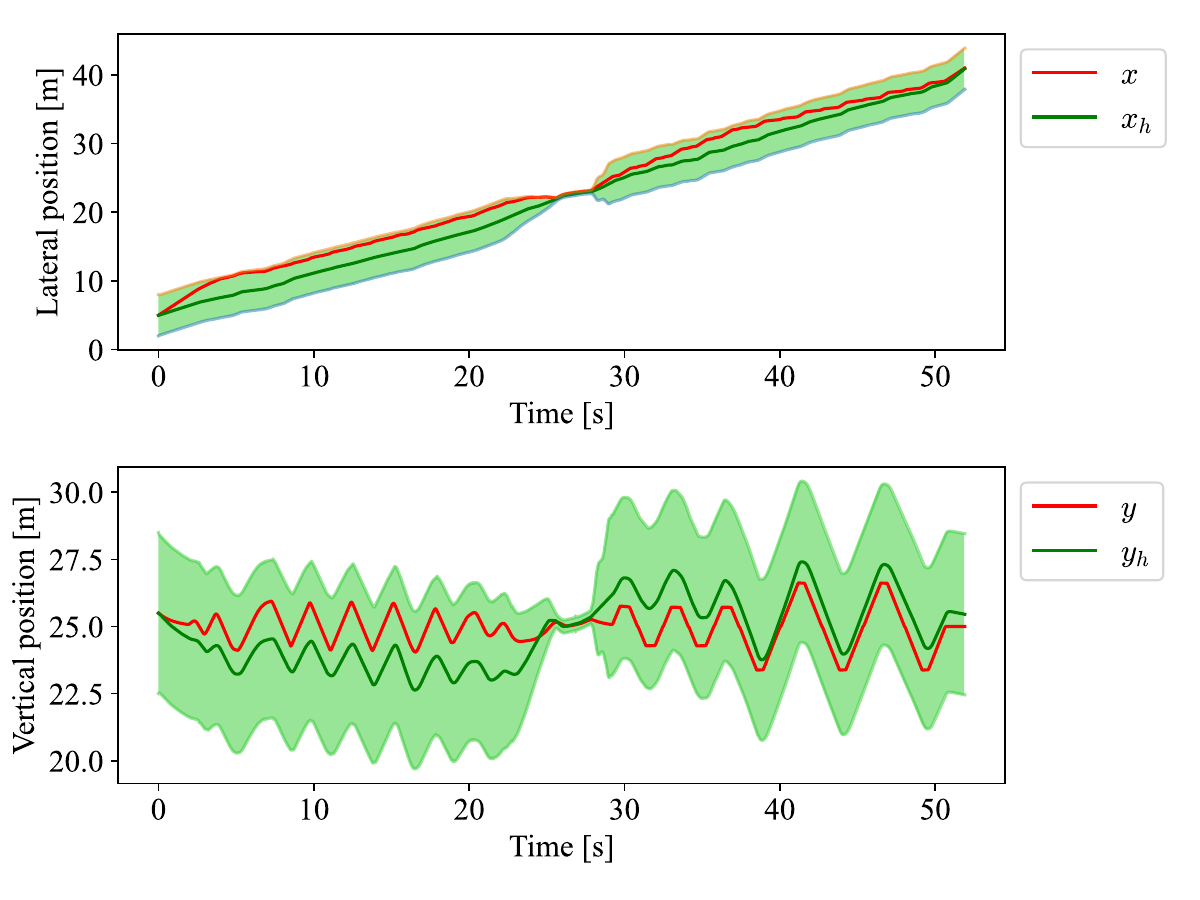}
\vspace{-8mm}
\caption{Lateral and vertical positions of a representative operator over time using the proposed method. The red line represents the vehicle position, and the green line represents the center of the acceptable region.}
\vspace{5mm}
\label{fig:5}
\end{figure}

\begin{figure}[t]
\centering
\includegraphics[width=\linewidth]{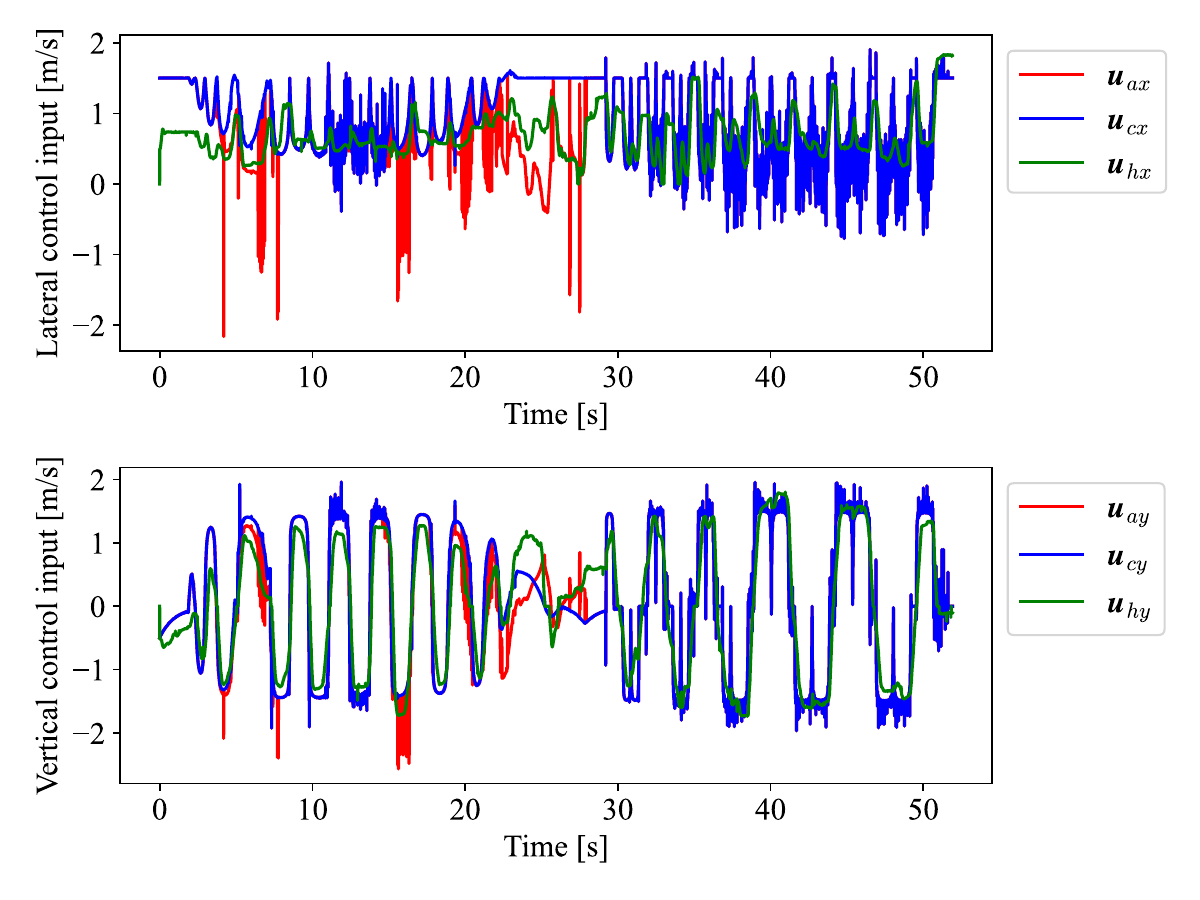}
\vspace{-8mm}
\caption{Control inputs of a representative operator using the proposed method. $\bm{u}_a$ denotes the input to the vehicle, $\bm{u}_c$ denotes the F-AC input, and $\bm{u}_h$ denotes the HSC input.}
\vspace{5mm}
\label{fig:6}
\end{figure}

Representative simulation results are shown in Figs. \ref{fig:4}–\ref{fig:6}.
As shown in Fig. \ref{fig:4}, the simple HSC method results in noticeable tracking errors along the entire path. In contrast, the proposed method demonstrates improved performance: while minor errors are observed in Area A, the system follows the reference path with high accuracy in Area C. However, in Area B—where the vehicle must stop—a significant deviation from the reference trajectory is observed. This is likely due to unintended joystick movement caused by tight gripping of the gripper to stop the vehicle.
As shown in Fig. \ref{fig:5}, the acceptable region defined by $\bm{x}_h$ and $\beta$, along with the system position $\bm{x}_a$. The upper figure shows the lateral (left–right) direction, while the lower figure represents the vertical (up–down) direction. The system position remains within the acceptable region throughout the trial. Even when $\beta$ decreases around 27–29 seconds, the system state stays close to the boundary without violating the acceptable region constraint.
As shown in Fig. \ref{fig:6}, the proposed method closely approximates F-AC in Area C. In contrast, during the period from 22 to 29 seconds, the system generates control inputs that closely follow the human’s intention, effectively switching from F-AC to human control.

The table \ref{tab:1} shows the results of root mean square error (RMSE) averages for the four participants and the average time required to achieve path following.
In Areas A and C, which require smooth and agile maneuvers, the proposed method achieved higher accuracy and shorter tracking times, demonstrating its effectiveness. In contrast, in Area B, the RMSE was larger for the proposed method, indicating greater deviation from the path. This deviation likely resulted from operators moving the joystick while gripping the gripper to intervene in the input. Consequently, the RMSE for the entire path was slightly higher for the proposed method; however, the overall completion time was reduced.

\begin{table}[]
\caption{Average RMSE and completion time.}
\label{tab:1}
\begin{tabular}{|c|cc|cc|}
\hline
         & \multicolumn{2}{c|}{RMSE   {[}m{]}}        & \multicolumn{2}{c|}{Required time {[}s{]}} \\ \hline
         & \multicolumn{1}{c|}{Proposed} & Simple HSC & \multicolumn{1}{c|}{Proposed} & Simple HSC \\ \hline
Area A   & \multicolumn{1}{c|}{0.103}    & 0.149      & \multicolumn{1}{c|}{19.634}   & 22.582     \\
Area B   & \multicolumn{1}{c|}{0.398}    & 0.154      & \multicolumn{1}{c|}{7.658}    & 7.682      \\
Area C   & \multicolumn{1}{c|}{0.079}    & 0.161      & \multicolumn{1}{c|}{24.042}   & 26.870     \\ \hline
All Areas & \multicolumn{1}{c|}{0.180}    & 0.159      & \multicolumn{1}{c|}{51.333}   & 57.134     \\ \hline
\end{tabular}
\vspace{5mm}
\end{table}

\section{Discussion}
The experimental results as shown in Table. \ref{tab:1} demonstrate the effectiveness of the proposed CBF-based, human--centered framework in smooth coupling F-AC and HSC across a path-following scenario. In contrast to conventional methods that rely on force-based modulation, our approach allows disregarding interface inputs according to human intent.
Enabling the operator to specify control boundaries rather than issuing continuous direct commands represents a more intuitive and interpretable control paradigm. This human--centered formulation enhances safety and stability, especially in teleoperation contexts characterized by uncertainty, delayed feedback, or the operator's workload. 

In Fig. \ref{fig:4}, Area B indicates that although the tracking error increased, it remained within the acceptable region. This implies that the rise in tracking error may have occurred because the operator could not accurately perceive the center position within that region.
The structure and control characteristics of the joystick, including the gripper affect the operator to perceive HSC guidance.
As the method approached fully manual operation, joystick movement naturally became more critical. However, gripping the gripper was necessary to reach fully manual operation, and the resulting increase in arm impedance may have reduced the effectiveness of HSC assistance.
We expect that this issue can be mitigated by introducing LoHA and adjusting the intensity of HSC according to the gripper’s angle.

\section{Conclusion}
This study presented a human–centric cooperative control framework that smoothly couples F-AC and HSC via control barrier functions. The framework addresses two key challenges: (1) maximizing the use of F-AC by allowing the human operator to dynamically define the F-AC region via CBF approach through a dedicated interface, and (2) facilitating timely human intervention by smoothly coupling F-AC and HSC outside this region. By guaranteeing operation within the operator-defined region in F-AC mode, the proposed method aligns machine autonomy with human intent while preserving the ability for the human to intervene whenever necessary.
A human-in-the-loop pilot experiment in a virtual teleoperation environment demonstrated that the proposed method outperforms conventional haptic shared control, particularly in regions requiring high-precision tracking and timely human intervention. The system successfully maintained the state within the operator-defined region, and adapted appropriately to changes in task demands. The experiments also revealed that, during transitions to fully manual control, unintended inputs were generated due to additional forces exerted on the input device.

As this study conducted only a pilot experiment, the next step is to perform full-scale experiments to rigorously verify the functionality and effectiveness of the proposed method. Another important direction of the future study is to identify the causes of the observed unintended actions and develop countermeasures. We hypothesize that these actions originate from the human operator's insufficient perception of the guidance force of the HSC.
To address this, consider combining partially variable LoHA, in which LoHA is adjusted separately for each axis of the joystick, with human intention estimation.
By estimating human intention for each axis of the joystick and setting the LoHA value for each, direction-specific assistance can be provided according to the task.

\bibliography{ifacconf}
\end{document}